\documentclass{article}
\usepackage{graphicx} 
\usepackage{amssymb}
\usepackage{amsmath}
\usepackage{lineno}
\newcommand{\comment}[1]{}
\usepackage{graphicx}%
\usepackage{multirow}
\usepackage{multicol}
\usepackage{amssymb,amsfonts}%
\usepackage[linesnumbered,ruled,vlined]{algorithm2e}
\usepackage{amsmath}
\usepackage{amsthm}%
\usepackage{mathrsfs}%
\usepackage[title]{appendix}%
\usepackage{textcomp}%
\usepackage{manyfoot}%
\usepackage{booktabs}%
\usepackage{float}
\usepackage{algorithmicx}%
\usepackage{algcompatible}
\usepackage{algpseudocode}%
\usepackage{listings}%
\usepackage[section]{placeins}
\usepackage{tabularx}
\usepackage[table]{xcolor}

\usepackage{makecell}

\usepackage[numbers,sort&compress]{natbib}

\usepackage{url}
\usepackage{hyperref}
\usepackage{authblk}

\title{Automatic camera orientation estimation for a partially calibrated camera above a plane with a line at known planar distance}
\author[1]{Gergely Dinya}
\author[1,2]{Anna Gelencsér-Horváth\thanks{Corresponding author: gha@inf.elte.hu}}

\date{}

\affil[1]{Faculty of Informatics, Eötvös Loránd University, Pázmány Péter stny. 1/C, 1117 Budapest, Hungary}
\affil[2]{Faculty of Information Technology and Bionics, Pázmány Péter Catholic University, Práter utca 50/a, 1083 Budapest, Hungary}

\begin{document}

\maketitle

\section{Background} 
\subsection{Camera projection}
\label{subsection:camera_projection}
Most modern cameras use a combination of lenses to direct light rays reflected off the surface of an object onto a sensor plate to capture an image. 
Mathematically, this process can be described using the intrinsic and the extrinsic matrices and a set of distortion coefficients ~\citep{BurgerCalibration2016}.

The $3\times3$ intrinsic matrix  
\begin{equation}\label{eq:intrinsic-matrix}
    \begin{bmatrix}
    f_x & s_\theta & c_x \\
    0 & f_y & c_y \\
    0 & 0 & 1
    \end{bmatrix}
\end{equation}
contains the camera's inner properties and describes how to map 3D points in the camera coordinate system to pixel coordinates on the image plane. 
Focal length is defined as the distance between the camera lens's focal point and the image sensor; the corresponding parameters $(f_x,f_y)$ are adjusted by a scalar factor to represent the ratio between the pixels' width and height. 
The camera's optical axis intersects the sensor at the image centre, denoted by $(c_x,c_y)$. 
While skew $(s_\theta)$ describes the diagonal distortion (though this is negligible in most modern cameras).

The $3\times4$ extrinsic matrix
\begin{equation}\label{eq:2}
    \begin{bmatrix}
        r_{11} & r_{12} & r_{13} & t_1 \\
        r_{21} & r_{22} & r_{23} & t_2 \\
        r_{31} & r_{32} & r_{33} & t_3
    \end{bmatrix}
\end{equation}
contains information on the camera's orientation and position in the world $[Rotation|Translation]$. It describes the transformation from the world coordinate system to the camera coordinate system.

The distortion coefficients are needed because lenses introduce geometric distortion into the image; this results in pixels being displaced from their true position by $(\delta x,\delta y)$.
Lens distortions can be grouped into three categories ~\citep{ronda2018geometrical} radial, tangential (also known as decentering) and thin-prism distortions.
The radial distortion is a result of the variation in light refraction inside the lens, which primarily affects wide-angle (\emph{fisheye}) lenses and manifests in the displacement of pixels along radial lines centered around the optical axis.
A misalignment between the center of the lens and the image sensor is the cause of the tangential distortion.
While the thin-prism distortion is a result of a tilt in the angle of the image sensor.
The number of coefficients can vary based on the types of distortions one has to deal with and the degree of precision of the representing equations.
To express the distortion, we use~\citep{opencv2008gary} the Brown-Conrady distortion model~\citep{brown1971close}, with  Eq. \ref{eq:3} and Eq. \ref{eq:4} to express the offset along the \emph{x} and \emph{y} axis, respectively. 
\begin{equation}\label{eq:3}
\begin{split}
    \delta x = &(x-c_x)(1 + k_1 \cdot r^2 + k_2 \cdot r^4 + k_3 \cdot r^6) + \\
    &\left[p_1 \cdot (r^2 + 2(x-c_x)^2) + 2p_2 \cdot (x-c_x) \cdot (y - c_y)\right]
\end{split}
\end{equation}
\begin{equation}\label{eq:4}
\begin{split}
    \delta y = &(y-c_y)(1 + k_1 \cdot r^2 + k_2 \cdot r^4 + k_3 \cdot r^6) + \\
    &\left[ 2p_1 \cdot (x-c_x) \cdot (y - c_y) + p_2 \cdot (r^2 + 2(y-c_y)^2)\right]
\end{split}
\end{equation}
The projection of a 3D coordinate onto the 2D image can be expressed by the following equations~\citep{BurgerCalibration2016}, where \emph{distort} is a 
\begin{math}
    \mathbb{R}^2\rightarrow\mathbb{R}^2
\end{math}
function, that offsets the undistorted $x$, $y$ coordinates by Eq.~\ref{eq:3} and Eq.~\ref{eq:4} respectively.
\begin{equation}
    \begin{bmatrix}
        x^{'}\\
        y^{'}\\
        z^{'}
    \end{bmatrix}
    =
    \begin{bmatrix}
    f_x & s_\theta & c_x \\
    0 & f_y & c_y \\
    0 & 0 & 1
    \end{bmatrix}
    \times
    \begin{bmatrix}
        r_{11} & r_{12} & r_{13} & t_1 \\
        r_{21} & r_{22} & r_{23} & t_2 \\
        r_{31} & r_{32} & r_{33} & t_3
    \end{bmatrix}
    \times
    \begin{bmatrix}
        X\\
        Y\\
        Z\\
        1
    \end{bmatrix}
\end{equation}
\begin{equation}
    \begin{bmatrix}
        x^{''}\\
        y^{''}
    \end{bmatrix}
    =
    \begin{bmatrix}
        x^{'}/z^{'}\\
        y^{'}/z^{'}
    \end{bmatrix}
\end{equation}
\begin{equation}
    \begin{bmatrix}
        u\\
        v
    \end{bmatrix}
    =
    distort(x^{''}, y^{''})
\end{equation}

\section*{Orientation estimation for partially calibrated cameras}
\textbf{Orientation estimation}

\begin{figure}[H]
    \centering
    \includegraphics[width=0.8\linewidth]{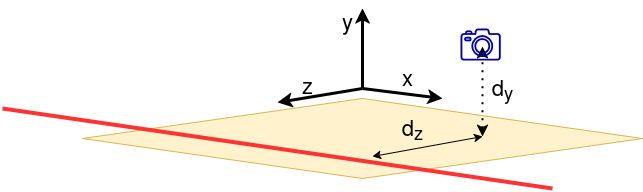}
    \caption{Camera position in the scene. Distances $d_z$ and $d_x$ are assumed as known}
    \label{fig:scene_setup}
\end{figure}

To compute camera orientation 
we need to estimate roll, pitch, and yaw.  
In the following calculations, we define the height from the plane or surface that is observed as a value along the y-axis, the distance toward the opposite side of the aviary (width) as movement along the z-axis, and the x-axis as the horizontal direction, as illustrated in Fig.~\ref{fig:scene_setup}.
We aim to calculate the extrinsic parameters along the three axes. 

We assume the camera height as fixed and known. %
We require a straight reference line at a fixed and known z-axis distance from the camera. 
This line can be determined from landmarks and task-specific logic. For example, from the intersection of the wall and floor in a room, or from the edges of objects with known dimensions, which can be detected automatically using segmentation or edge detection models.
With these constraints, we can determine both the roll and the pitch.

Roll is estimated by selecting the two extreme pixels where the detected 
line
meets the image borders and ensuring their inverse-projected z-coordinates are as close as possible, as defined in Eq.\ref{eq:roll_calc}, ensuring alignment between the projected surface plane and the detected reference line. 

\begin{equation}\label{eq:roll_calc}
        \lambda =  tan^{-1}\left(\frac{ y'_1 -  y'_2}{ x'_1 -  x'_2}\right)
\end{equation}
In this equation, ($x_1$,$y_1$) and ($x_2$,$y_2$) are the pixel coordinates of the two extreme edge points, ($c_{x1}$,$c_{y1}$) and ($c_{x2}$,$c_{y2}$) are the principal points, and $f_{x1},f_{y1},f_{x2},f_{y2}$ are the respective focal lengths along the $x$ and $y$ axes.

To estimate pitch, we use the central pixel of the reference line $(x_0,y_0)$, as it remains unaffected by roll. 
Pitch is computed using Eq.\ref{eq:pitch_calc},

\begin{equation}\label{eq:pitch_calc}
    \theta = tan^{-1}\left(\frac{c_0 - z_0\cdot\frac{y_0 - c_y}{f_y}}{(z_0 + c_0\cdot\frac{y_0 - c_y}{f_y})}\right)
\end{equation}

\noindent
where $c_0$ represents the camera height above the observed plane, $z_0$ is the known z-distance of the reference line and the 
point on the plane vertically aligned with the camera position
$y_0$ is the y-coordinate of the central pixel of the detected reference line, 
$c_y$ is the principal point in the y-direction, $f_y$ is the focal length in the y-direction. 
The first two elements on the main diagonal of the intrinsic camera matrix $K$ are $f_x$,  and $f_y$, while $c_x$, $c_y$ can be found in the last column, in the first and second rows, respectively, by definition.

As for yaw, 
estimation 
needs a line perpendicular to the aviary edge, across the plane or surface that is observed. 


\section*{Derivation of the formulas for the orientation estimation}
\label{sec:appendix_a}
Given a camera's intrinsic parameters and a plane located at a set distance $c_0$ below the camera, we want to find a camera orientation where the predicted plane 
fits to 
with the perceived plane (plane or surface that is observed) so that we can accurately predict the positions of 
objects 
on the plane.

We denote the $K$ intrinsic matrix of the camera, the $c_0$ distance along the y axis and a perceived line
and its distance along the z-axis.
First, we find the roll value for which the point of the perceived line in the middle of the image (no relative translation along the x-axis) is estimated to be $z_0$ distance away from the camera along the z-axis. 
Afterwards, we need to find a pitch value so that every point on the line has the same predicted distance along the z-axis.


\textbf{Roll}
First, let's define the inverse-projection that takes a 2D-pixel coordinate, the camera's intrinsic matrix, rotation vector and the distance of the constraining $xz-plane$ and returns the coordinates obtained by intersecting the homogeneous ray going through the pixel with the constraining plane.
\begin{equation}
    ir\_ray = 
    R^{-1}
    \times
    \begin{bmatrix}
        x'\\
        y'\\
        1
    \end{bmatrix}
        =
    R^{-1}
    \times
    \begin{bmatrix}
        \frac{x - c_x}{f_x}\\
        \frac{y - c_y}{f_y}\\
        1
    \end{bmatrix}
        =
    \begin{bmatrix}
        x''\\
        y''\\
        z''
    \end{bmatrix}
\end{equation}
\begin{equation}
    p\_constraint = c_0 / y''
\end{equation}
\begin{equation}
    relative\_point\_on\_the\_plane = 
    \begin{bmatrix}
        x'' \cdot p\_constraint\\
        p\_constraint\\
        z'' \cdot p\_constraint
    \end{bmatrix}
    =
    \begin{bmatrix}
        X\\
        Y\\
        Z
    \end{bmatrix}
\end{equation}

We want to find a specific rotation ($\theta$) along the x-axis for which the z-coordinate of the inverse projection of the pixel in the middle of the perceived line (thus not affected by the roll) is $z_0$.

Let's modify the original equations by replacing the 3D rotation matrix $R$ with $R_x$ that only deals with rotation along the x-axis.
\begin{equation}
    \begin{bmatrix}
        x''\\
        y''\\
        z''
    \end{bmatrix}
    =
    \begin{bmatrix}
        1&0&0\\
        0&cos\theta&sin\theta\\
        0&-sin\theta&cos\theta
    \end{bmatrix}
    \times
    \begin{bmatrix}
        x'\\
        y'\\
        1
    \end{bmatrix}
    =
    \begin{bmatrix}
        x'\\
        cos\theta\cdot y'+sin\theta\\
        cos\theta - sin\theta\cdot y'
    \end{bmatrix}
\end{equation}
The relative point on the plane becomes:
\begin{equation}
    \begin{bmatrix}
        X\\
        Y\\
        Z
    \end{bmatrix}
    =
    \begin{bmatrix}
        \frac{c_0\cdot x''}{y''}\\
        c_0\\
       \frac{c_0\cdot z''}{y''}
    \end{bmatrix}
\end{equation}

We want to find $\theta$ for which $Z=z_0$.

\begin{equation}
    z_0 = z'' \cdot \frac{c_0}{y''}
\end{equation}
\begin{equation}
    z_0\cdot cos\theta\cdot y'+z_0\cdot sin\theta = c_0\cdot cos\theta - c_0\cdot sin\theta\cdot y'
\end{equation}
\begin{equation}
    z_0\cdot y'+z_0\cdot tan\theta = c_0 - c_0\cdot tan\theta\cdot y'
\end{equation}
\begin{equation}
    tan\theta = \frac{c_0 - z_0\cdot y'}{z_0 + c_0\cdot y'}
\end{equation}
\begin{equation}
    \theta = tan^{-1}\left(\frac{c_0 - z_0\cdot y'}{z_0 + c_0\cdot y'}\right)
\end{equation}

\textbf{A.3}
Given the $\theta$ rotation along the x-axis, we want to find a rotation ($\lambda$) along the z-axis, for which any two points on the perceived line will have the same z coordinate. As we will see, this is essentially just the angle of the line.

Let's modify the original equations by replacing the 3D rotation matrix $R$ with $R_{xz}$ that only deals with rotations along the x and z axes.

\begin{equation}
\begin{split}
    \begin{bmatrix}
        x'''\\
        y'''\\
        z'''
    \end{bmatrix}
    =
    \begin{bmatrix}
        cos\lambda&sin\lambda&0\\
        -sin\lambda cos\theta&cos\lambda cos\theta&sin\theta\\
         sin\lambda sin\theta&-cos\lambda sin\theta&cos\theta
    \end{bmatrix}
    \times
    \begin{bmatrix}
         x'\\
         y'\\
        1
    \end{bmatrix}
    = \\
    \begin{bmatrix}
        cos\lambda \cdot  x' + sin\lambda\cdot y'\\
        -sin\lambda \cdot cos\theta\cdot x'+cos\lambda \cdot cos\theta\cdot y' +sin\theta\\
        sin\lambda \cdot sin\theta\cdot x'-cos\lambda \cdot sin\theta \cdot  y' + cos\theta
    \end{bmatrix}
\end{split}
\end{equation}

The relative point on the plane becomes.
\begin{equation}
    \begin{bmatrix}
        X\\
        Y\\
        Z
    \end{bmatrix}
    =
    \begin{bmatrix}
         \frac{c_0\cdot x'''}{y'''}\\
        c_0\\
        \frac{c_0\cdot z}{y'''}
    \end{bmatrix}
\end{equation}

Let

\begin{equation}
        c_z = Z \cdot cos\theta +  c_0\cdot sin\theta
\end{equation}

We can rearange the equation for Z as follows:

\begin{equation}
        sin\lambda \cdot (c_z\cdot x') - cos\lambda \cdot (c_z\cdot y') =
        Z\cdot sin\theta - c_0 \cdot cos\theta
\end{equation}

We want to find a $\lambda$ for which two points on the line $(x_1,y_1)$,$(x_2,y_2)$ will have the same $Z$ coordinate, satisfying the following equation:
\begin{equation}
        sin\lambda \cdot (c_z\cdot x'_1) - cos\lambda \cdot (c_z\cdot y'_1) = 
        sin\lambda \cdot (c_z\cdot x'_2) - cos\lambda \cdot (c_z\cdot y'_2)
\end{equation}
\begin{equation}
        sin\lambda \cdot (c_z\cdot x'_1 - c_z\cdot x'_2) =
        cos\lambda \cdot (c_z\cdot y'_1 - c_z\cdot y'_2)
\end{equation}
\begin{equation}
        tan\lambda =  \frac{c_z\cdot (y'_1 - y'_2)}{c_z\cdot(x'_1 -  x'_2)}
\end{equation}
\begin{equation}
        \lambda =  tan^{-1}\left(\frac{ y'_1 -  y'_2}{ x'_1 -  x'_2}\right)
\end{equation}

This is essentially just the angle of the perceived line.

\bibliographystyle{unsrtnat}
\bibliography{main}  

\end{document}